\documentclass{pprai}

\usepackage[utf8]{inputenc}
\usepackage[T1]{fontenc}
\usepackage{graphicx}
\usepackage{amsthm}
\usepackage{txfonts}
\usepackage{url}
\usepackage{algpseudocode}
\usepackage{xcolor}
\usepackage{hyperref}
\usepackage{listings}
\usepackage{subcaption}

\title{Interpreting and learning voice commands with a Large Language Model for a robot system}
\headtitle{Interpreting and learning voice commands\dots for a robot system}

\author{Stanislau Stankevich$^{1[0009-0007-2432-4500]}$, Wojciech Dudek$^{1[0000-0001-5326-1034]}$}
\headauthor{S. Stankevich, W. Dudek}
\affiliation{%
  $^1$Warsaw University of Technology\\
 Faculty of Electronics and Information Technology\\
  Nowowiejska 15/19, 00-662 Warsaw, Poland\\
  wojciech.dudek@pw.edu.pl\\
}


\begin{document}
\maketitle

\begin{abstract}
Robots are increasingly common in both industry and daily life, such as in nursing homes where they can assist staff. A key challenge is developing intuitive interfaces for easy communication. The use of Large Language Models (LLMs) like GPT-4 has enhanced robot capabilities, allowing for real-time interaction and decision-making. This integration improves robots' adaptability and functionality. This project focuses on merging LLMs with databases to improve decision-making and enable knowledge acquisition for the request interpretation problems.
\end{abstract}



\section{Introduction}

We observe the necessity for user-friendly interfaces for seamless human-robot interaction, with voice interfaces being a prime example \cite{10.1145/3386867}. The use of Large Language Models (LLMs) like GPT-4 has enhanced robot capabilities, allowing for real-time interaction and decision-making, as demonstrated in \cite{Klingensmith2023RobotsChat}.

\begin{figure}[htb]\centering
  \begin{minipage}[b]{0.6\textwidth}
    \centering \includegraphics[width=\linewidth]{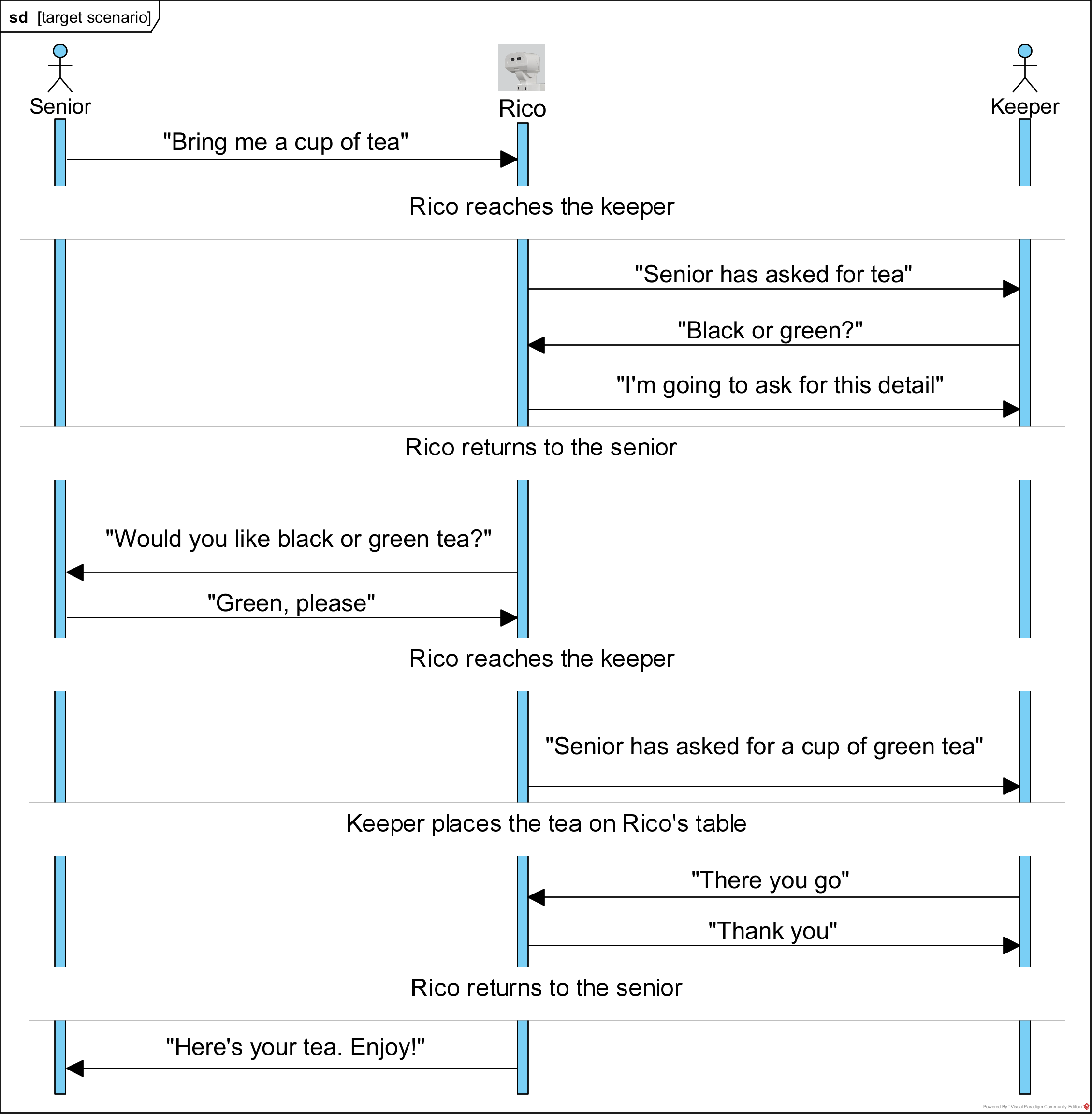}
    \caption{Scenario with bringing an item}
    \label{fig:seq-bring-goods}
  \end{minipage}
  \hskip 0.5 cm
  \begin{minipage}[b]{0.3\textwidth}\centering
        \includegraphics[width=0.8\linewidth]{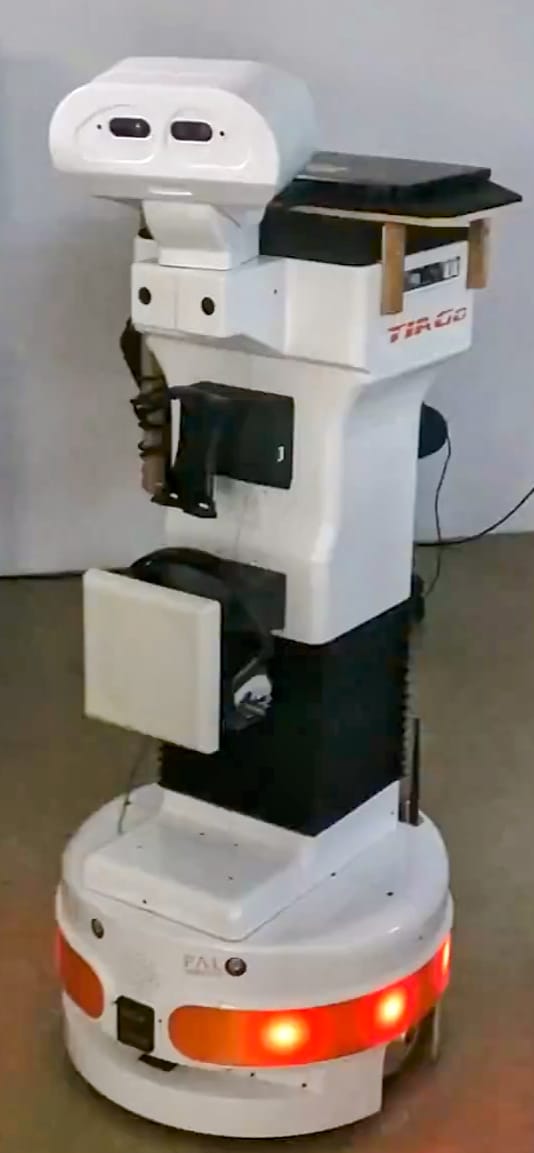}
        \caption{Rico robot}
        \label{fig:rico}
  \end{minipage}
  
\end{figure}

This work seeks to enhance the capabilities of Rico (Fig.~\ref{fig:rico}), a robot developed using PAL Robotics' TIAGo platform \cite{Pags2016TIAGoTM,intent-tiago}, designed to assist seniors in care homes. Its example application involves two main human roles: the Senior, an elderly individual who may face mobility or other challenges, and the Keeper, a staff member responsible for assisting the seniors. Rico, the robot, serves as a third participant, designed to support the Keeper by taking on simpler tasks, e.g. delivering goods to the seniors.


The scenario is an item transportation from the keeper to the senior at the senior's request. However, the item may have various attributes, like tea's temperature, volume, and additional ingredients. The robot must not only fulfil these requests but also learn the available options (like only black coffee being available) by interpreting dialogues.

The process of identifying the user's purpose behind a request is known in NLP as \textit{intent detection}. Additional details included in a request are called \textit{parameters} or \textit{slots} of the intent. For instance, the intent "bring coffee" might include a "size" parameter, while "sing a song" could have parameters like "song name" and "volume". The process of extracting the values of the parameters from user's request is called \textit{slot filling}.

The aim of this work is to enable the system to dynamically expand it's knowledge about given task by identifying and incorporating new intents and slots the during conversations.

The sequence diagram in Fig. \ref{fig:seq-bring-goods} presents this use case scenario. The subsequent request for tea should include an immediate tea-type clarification question for the Senior. This learning behaviour should be continued.

\section{The conversation system}
\begin{figure}[htb]
    \centering \includegraphics[width=0.9\linewidth]{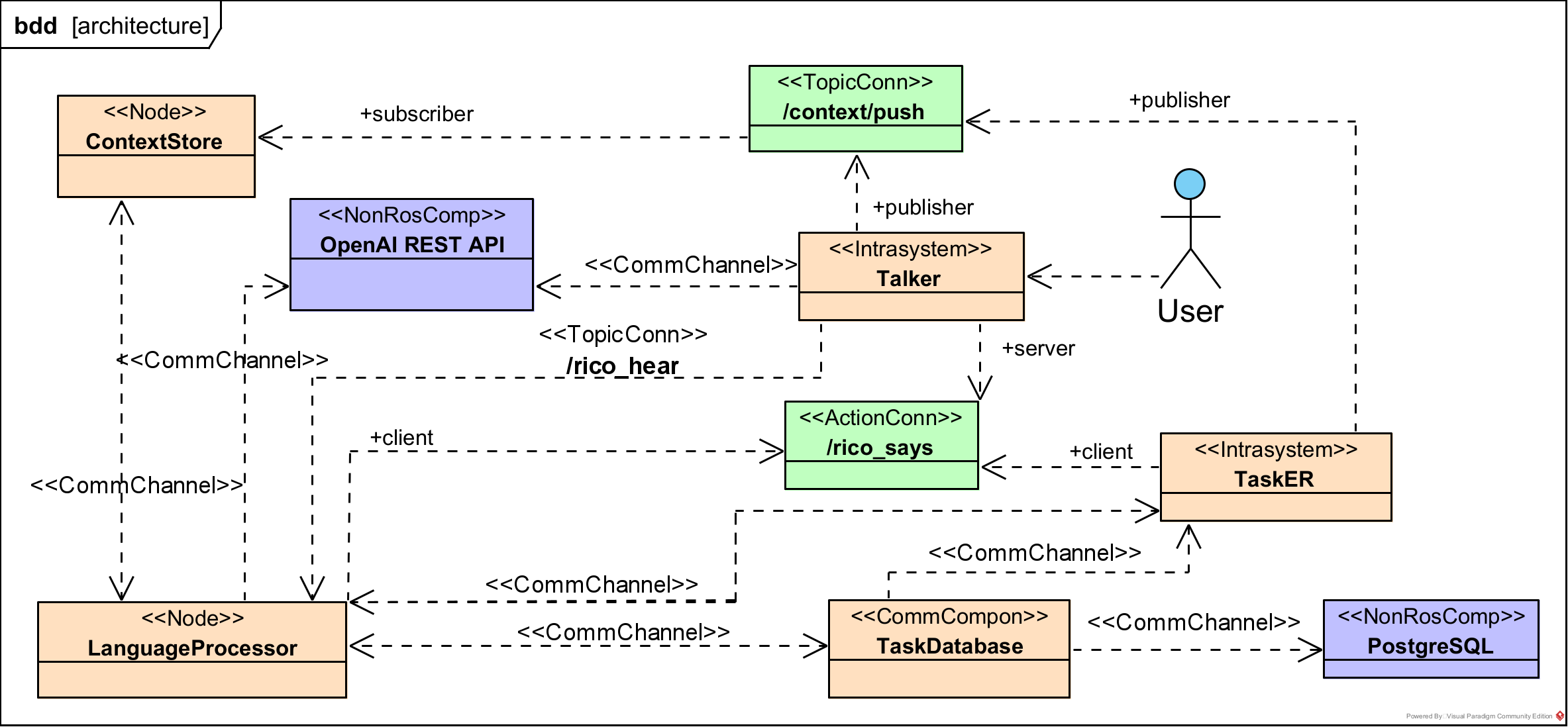}
    \caption{System architecture}
    \label{fig:final}
\end{figure}

The system has a~ROS-based architecture~\cite{quigley2009ros}. It is modelled with MeROS \cite{meros} and it is shown in Fig.~\ref{fig:final}. Each system's component serves a specific function within the system.

User interactions with the robot are handled by the ROS node \texttt{Talker}. It converts speech to text and text to speech. This involves turning audio requests into text and the robot's responses into audible speech using OpenAI's models with the web API. The text-converted requests are sent to the LanguageProcessor (\texttt{LangProc}) component for further processing.

\begin{figure}[htb]
    \centering \includegraphics[width=0.9\linewidth]{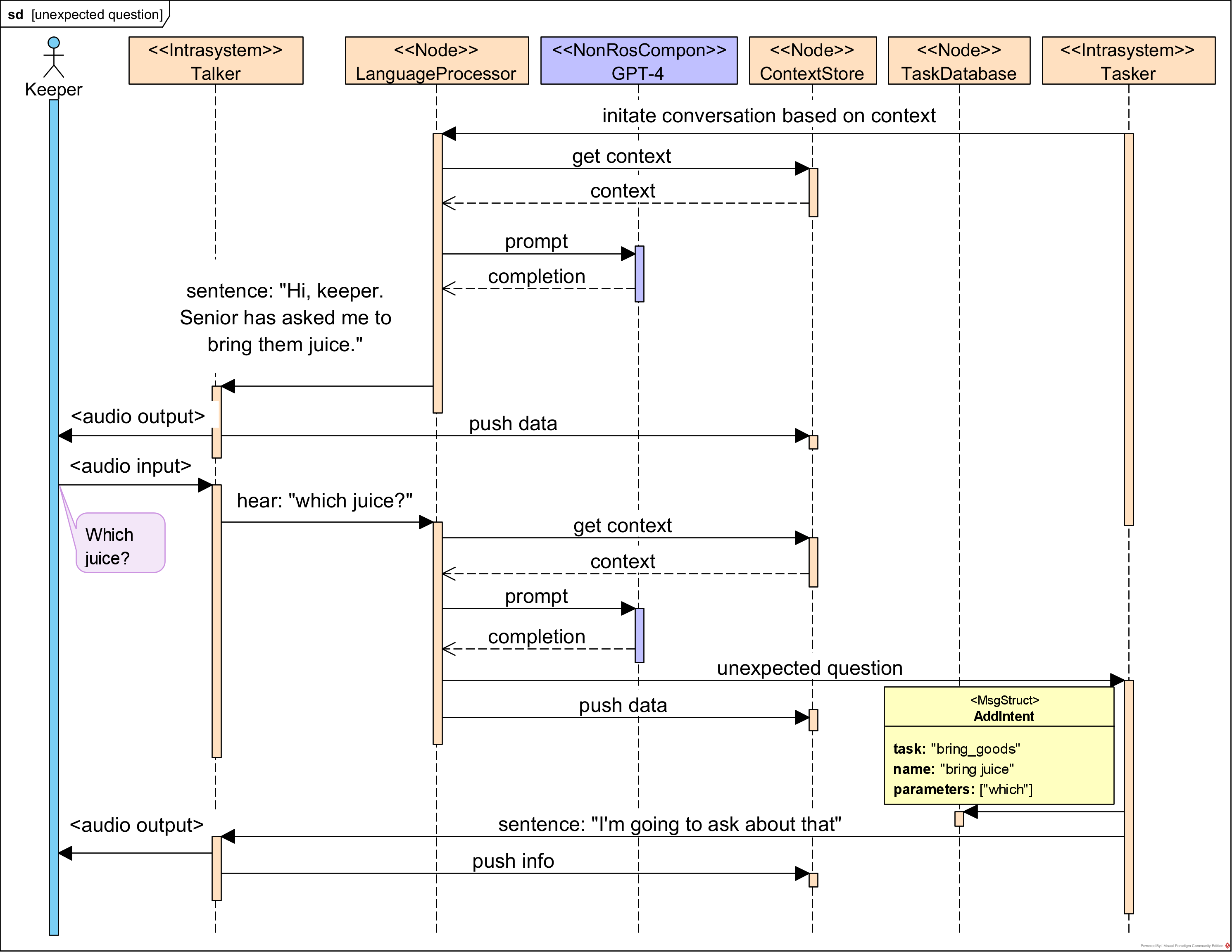}
    \caption{Modules interaction - unexpected question}
    \label{fig:Solution with GPT-3.5 - unexpected question}
\end{figure}

The ROS node ContextStore (\texttt{CS}) records and organises event data during user sessions, like "Rico heard 'bring tea'" or "Rico completed a task." It enables the system to process this information easily, e.g. joining it to prompts for LLM, thus enhancing user interaction.

TaskER is a framework for task management in ROS-based mobile \cite{Schedulingwithtasker} and mobile manipulation \cite{Dudek:2021_phd-twiki} robots. It enables task delegation to a robot by sending a message on a specific ROS topic. the TaskER module encompasses a task repository represented by state machines, defined with SMACH \cite{smach}.


TaskDatabase (\texttt{TD}) is a ROS interface to a PostgreSQL database. It stores known intents along with their parameters and a mapping between intents and tasks that they trigger.

The \texttt{LangProc} handles natural language processing in the system using its connection to the LLM (OpenAI GPT-4 API). It crafts prompts for system tasks. For these prompts construction, the \texttt{LangProc} queries a database storing known user intents, gets the textual transcription of the user's request, and acquires the current context and scenario (e.g. who is the robot talking with and what was already mentioned in the conversation). Six prompt templates\footnote{\url{https://github.com/RCPRG-ros-pkg/rico-language-processor/tree/main/src/stories}} were found to be sufficient for achieving for achieving desired behaviour.

A request for an item (e.g. juice) initiates the application. This request reaches the \texttt{LangProc} module, which gathers data about intents known to the system from \texttt{TD} and then prompts GPT to guess user intent based on the current state of the conversation. If the system identifies the intent, \texttt{LangProc} passes it to TaskER, which then checks \texttt{TD} for the related task name and executes the task.


When the robot reaches the kitchen, it asks the keeper for the juice, sending a prompt request to the \texttt{LangProc}. \texttt{LangProc} fetches the context, queries GPT-4, and the resulting text is vocalized by \texttt{Talker}. Throughout, \texttt{Talker} updates \texttt{CS} with the spoken sentences, extending the system's context.

The question "Which juice?" is set as unexpected by LLM. In consequence, the new intent "Bring juice" with parameter "which" is added to the database, and the robot returns to the senior for the juice type clarifying, as shown in Fig.~\ref{fig:Solution with GPT-3.5 - unexpected question}.

Subsequently, \texttt{LangProc} again queries GPT-4 with the conversation context, and LLM asks, "What kind of juice would you like?" due to detecting "Bring juice" intent with slot "which" not filled. \texttt{Talker} voices this question to the senior, who answers "Apple juice". This expands the context, and LLM extracts all the necessary information, which then reaches the TaskER and the scenario continues as before. The robot reaches the keeper, and GPT-4 generates the robot's initiating conversation with the keeper.

\section{Validation and Conclusion}

We executed the system in various "bring goods" scenarios, as demonstrated in the video\footnote{\url{https://vimeo.com/863071575}}. The system effectively conducted the primary scenario, adeptly responding to unexpected inquiries and remembering the "sugar" and "lemon" attributes of the "bring tea" task. However, there is a need to enhance the system's performance. Secondly, the system sometimes struggles with tasks with more than five parameters. We observed that GPT-4 generates inaccurate or "hallucinated" responses, though infrequently. E.g. "Bring me a cup of tea with sugar" may sometimes be misinterpreted, and the "blackOrGreen" slot may be filled as "Black".


In conclusion, we propose an adaptive voice-operated interface for robots. The challenge was integrating the LLM abilities with databases, state machines and the robot controller. Our system uses ROS, speech-to-text and text-to-speech models, and a relational database to deliver vocal conversation functionality with tasks and their parameters' interpretation and exploration. It is usable by a~waiter robot, though the system would need customization. Adjusting other Rico's tasks for dynamic intent knowledge expanding requires changing other TaskER state machines by integrating them with \texttt{LangProc} and \texttt{TD}. The future version could model the robot's environment with LLM prompts and plan tasks by querying LLM.

The most recognized current GPT application in robotics is the Spot robot controlled with voice commands \cite{Klingensmith2023RobotsChat}. Our system contrasts with it by extending robot's database using LLMs, instead of providing an API to the robot for LLM. While the idea of LLM-controlled robot gains attention, the robot's ability to explore its environment by using LLM is also desirable.



\section*{Acknowledgment}
Research was funded by the Centre for Priority Research Area Artificial Intelligence and Robotics of Warsaw University of Technology within the Excellence Initiative: Research University (IDUB) programme.
\bibliography{pprai}
\bibliographystyle{pprai}

\end{document}